\documentclass[letterpaper]{article} 
\usepackage{aaai24}  
\usepackage{times}  
\usepackage{helvet}  
\usepackage{courier}  
\usepackage[hyphens]{url}  
\usepackage{graphicx} 
\usepackage{natbib}  
\usepackage{caption} 
\usepackage{algorithm}
\usepackage{algorithmic}

\usepackage{caption}
\usepackage{subcaption}
\usepackage{amsmath}
\usepackage{amssymb}
\usepackage{mathtools}
\usepackage{amsthm}

\usepackage[capitalize,noabbrev]{cleveref}
\usepackage{enumitem}

\newcommand{\given}{\lvert}
\newcommand{\E}{\mathbb{E}}

\usepackage{newfloat}
\usepackage{listings}
\DeclareCaptionStyle{ruled}{labelfont=normalfont,labelsep=colon,strut=off} 
\floatstyle{ruled}
\newfloat{listing}{tb}{lst}{}
\floatname{listing}{Listing}
\title{Neural Amortized Inference for Nested Multi-agent Reasoning}
\author {
    Kunal Jha\textsuperscript{\rm 1},
    Tuan Anh Le\textsuperscript{\rm 2},
    Chuanyang Jin\textsuperscript{\rm 3},
    Yen-Ling Kuo\textsuperscript{\rm 4},\\
    Joshua B. Tenenbaum\textsuperscript{\rm 5},
    Tianmin Shu\textsuperscript{\rm 5,6}\\
}
\affiliations {
    \textsuperscript{\rm 1}Dartmouth College,
    \textsuperscript{\rm 2}Google Research, 
    \textsuperscript{\rm 3}New York University,
    \textsuperscript{\rm 4}University of Virginia,
    \textsuperscript{\rm 5}Massachusetts Institute of Technology,
    \textsuperscript{\rm 6}Johns Hopkins University\\
    kunal.a.jha.24@dartmouth.edu, tuananhl@google.com, cj2133@nyu.edu, ylkuo@virginia.edu, \{jbt, tshu\}@mit.edu
}

\usepackage{bibentry}

\begin{document}

\maketitle

\begin{abstract}
Multi-agent interactions, such as communication, teaching, and bluffing, often rely on higher-order social inference, i.e., understanding how others infer oneself. Such intricate reasoning can be effectively modeled through nested multi-agent reasoning. Nonetheless, the computational complexity escalates exponentially with each level of reasoning, posing a significant challenge. However, humans effortlessly perform complex social inferences as part of their daily lives. To bridge the gap between human-like inference capabilities and computational limitations, we propose a novel approach: leveraging neural networks to amortize high-order social inference, thereby expediting nested multi-agent reasoning. We evaluate our method in two challenging multi-agent interaction domains. The experimental results demonstrate that our method is computationally efficient while exhibiting minimal degradation in accuracy.
\end{abstract}



\section{Introduction}
 
\begin{figure}[t!]
\centering
\includegraphics[width=0.45\textwidth]{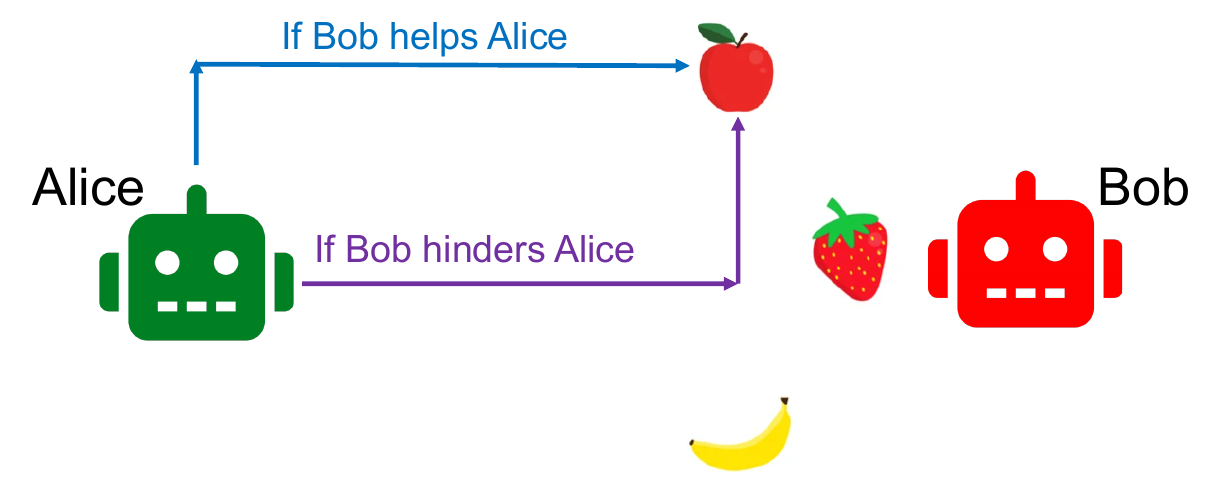}
    \caption{Illustration of nested reasoning between two agents. Alice wants to get the apple and Bob may help or hinder Alice. Agents do not know each other goals. It is crucial that Alice successfully infers Bob's intent so that Alice can choose the optimal path to either cooperate with Bob or compete with Bob optimally. Such inference requires nested reasoning -- in this case, Alice infers how Bob infers Alice.
    }
    \label{fig:intro}
\end{figure}

We reason about and act in a world that contains not only inanimate objects but also agents. Reasoning about other agents is key to our everyday abilities to empathize, communicate, teach, and more. Imbuing AI systems with adequate social inference is key to enabling rich multi-agent interactions. For instance, as illustrated in Figure~\ref{fig:intro}, to successfully interact with another agent, one would need to infer not only the agent's goal but also how the other agent infers the goal of oneself. Current multi-agent systems still cannot conduct nested multi-agent reasoning efficiently and flexibly. 

There have been prior works attempting to build agents that can conduct nested reasoning with each other to achieve sophisticated interactions. Most notably, Interactive Partially Observable Markov Decision Making Process (I-POMDP) \citep{gmytrasiewicz2005framework} provides a framework for modeling multi-agent systems with nested reasoning. While the framework is general and can be applied to a broad range of domains, we still cannot equip autonomous agents with human-like nested reasoning to date. One of the main challenges is that inference under the I-POMDP framework is recursive and hence slow. In complex domains in which there are large state, action, and goal spaces, solving I-POMDP becomes computationally intractable. 

However, we can make nested reasoning effortlessly in our daily life despite the complexity of the inference. This work aims to develop an efficient inference method for nested multi-agent reasoning. For this, we follow the I-POMDP framework and propose to amortize inference at different levels of I-POMDP by training neural networks to sample data-driven proposals. This allows us to sample only a small number of hypotheses to cover the ground truth, which drastically decreases the overall computation without sacrificing accuracy. 

We demonstrate our approach in two complex domains requiring higher-order social reasoning. In the first domain, we consider a two-player game in 2D grid worlds, Construction, in which one agent (Bob) needs to help or hinder another agent (Alice) but the two agents do not know each other's goal. To successfully infer the goal of Bob, one would need nested goal inference. Additionally, we evaluate our approach in a driving domain, in which drivers must infer each other's mental states (including goals and beliefs about the physical states) recursively to avoid collision and to signal danger to other drivers in anticipation of a crash. To successfully predict drivers' actions, a model must conduct nested goal and belief inference. Our experimental results show that, in both domains, the amortized inference greatly reduces the total amount of compute while maintaining a similar inference accuracy compared to conventional approaches. It can also estimate the uncertainty in inference and generalizes well to unseen scenarios.


\section{Related Work}
\label{sec:related_work}

\textbf{Nested Multi-agent Reasoning.} There have been prior works formulating nested reasoning between multi-agents. The most general formulation is the interactive partially observable Markov decision process (I-POMDP), proposed by~\citet{gmytrasiewicz2005framework}. Additionally, there have been models that address special cases in nested multi-agent reasoning. For instance, cooperative inverse reinforcement learning (CIRL) models human-robot cooperation in which a user needs to infer how a robot assistant infers the user's reward \citep{hadfield2016cooperative}. \citet{frank2012predicting} model a type of nested reasoning between a speaker and a listener, pragmatic reasoning, with the Rational Speech Act (RSA) modeling framework. \citet{tejwani2022social} propose a Social MDP framework to model complex social interactions that rely on nested reasoning about the reward functions of other agents. While these are all powerful frameworks, there has not been much work on developing efficient inference algorithms for these frameworks in complex domains in which the hypothesis space on each level is very large. Conventional methods \citep{rathnasabapathy2006exact, doshi2009monte, seaman2018nested} can conduct explicit nested reasoning robustly but fail to scale to complex environments. Recently, \citet{han2019ipomdp} propose an end-to-end model to approximate the nested belief update as hidden state updates in a neural network to train multi-agent policies. However, such end-to-end models cannot conduct explicit nested reasoning with explainable goals and belief representations (e.g., they can generate actions given agents' goals but cannot infer agents' goals based on observed actions). They also cannot estimate uncertainty in inference. This work aims to fill in the gap by proposing an efficient inference method that is also explainable and robust.


\textbf{Theory of Mind Reasoning.} More broadly speaking, nested multi-agent reasoning is a type of Theory of Mind reasoning \citep{premack1978does}, in which agents need to infer each other's mental states based on observed actions. There are two main types of computational models for Theory of Mind: end-to-end methods based on neural networks such as Machine Theory of Mind network \citep{rabinowitz2018machine,chuang2020using} and model-based methods relying on generative models of agents such as Bayesian Theory of Mind \citep{baker2017rational, ullman2009help, shum2019theory, netanyahu2021phase}. Our approach combines both types of models to achieve fast (through neural networks) yet robust (through model-based reasoning) inference.

\textbf{Neural Amortized Inference.} There have been prior works on neural amortized inference for accelerating probabilistic inference in complex domains~\citep{le2017inference}, such as computer graphics \citep{ritchie2016neurally} and particle physics \citep{baydin2019etalumis}. These works have demonstrated that neural networks can be trained to sample data-driven proposals that are more likely to include the ground-truth hypothesis. Consequently, the inference algorithms would only need to maintain a small set of particles in order to achieve high accuracy. In this paper, we adopt the idea of neural amortized inference to accelerate the inference following I-POMDP formulation as a general solution for nested multi-agent reasoning.

\section{Background}


\subsection{I-POMDP}\label{sec:ipomdp}

\begin{figure}[t!]
    \centering
    \includegraphics[width=0.47\textwidth]{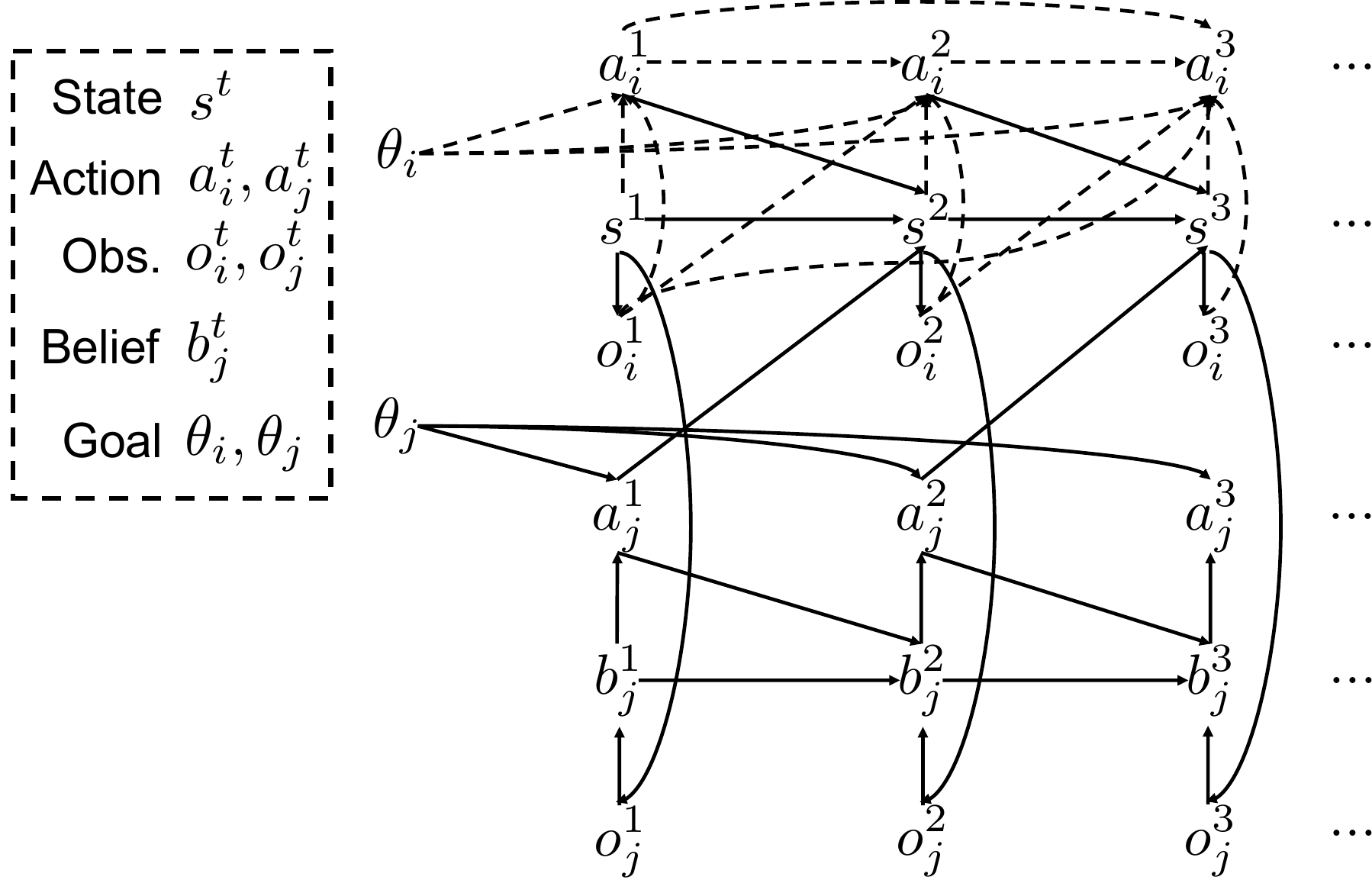}
    \caption{Graphical model for I-POMDP where an agent $i$ plans its actions while modeling another agent $j$. Dashed arrows describe the \emph{planner} that outputs the probabilities of $i$'s actions. The solid arrows describe the dynamics of the environment and another agent $j$ that the planner is based on.}
    \label{fig:ipomdp}
\end{figure}


Interactive Partially Observable Markov Decision Process (I-POMDP) extends POMDP by recursively modeling other agents in the environment~\citep{gmytrasiewicz2005framework}.
For rotational convenience and without loss of generality, we assume that there are two agents, $i$ and $j$. Agent $i$ is the ego agent, which models and interacts with agent $j$. 

As illustrated in \cref{fig:ipomdp}, an I-POMDP model states $s^{1:T}$\footnote{$T$ is the total number of steps.}, state observations of the two agents $o_i^{1:T}, o_j^{1:T}$, actions of the two agents $a_i^{1:T}, a_j^{1:T}$; for agent $j$, we additionally model its beliefs $b_j^{1:T}$ and other information about its mind that is relevant to its behavior, $\theta_j$. In this work, we define $\theta_i$ as agent $i$'s goal. But this can be extended to other types of information such as preferences.

Following \citet{doshi2009monte}, we inductively define $i$'s interactive state $is_{i, \ell}$ at level-$\ell$ as
\begin{itemize}[label={}]
    \item Level $0$: $is_{i, 0} = s$
    \item Level $1$: $is_{i, 1} = (s, b_{j, 0}, \theta_j)$ where $b_{j, 0}$ is a \emph{distribution} over $j$'s interactive state at level $0$, $is_{j, 0}$
    \item $\cdots$
    \item Level $\ell$: $is_{i, \ell} = (s, b_{j, \ell - 1}, \theta_j)$ where $b_{j, \ell - 1}$ is a \emph{distribution} over $j$'s interactive state at the previous level, $is_{j, \ell - 1}$ 
\end{itemize}

\begin{figure*}[t!]
     \centering
         \centering
         \includegraphics[width=\textwidth]{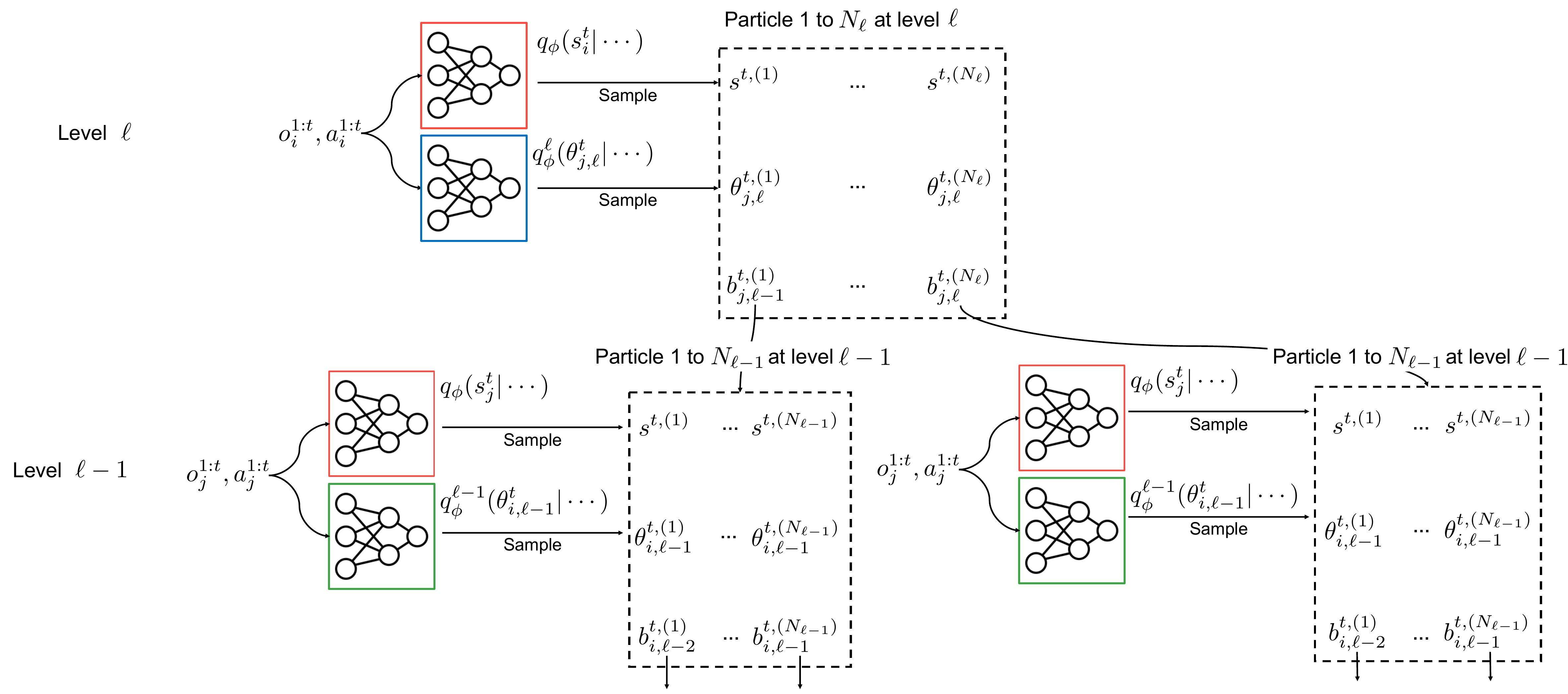}
        \caption{Illustration of particle sampling at each level using the recognition networks recursively. Networks with the same colors are the same.}
        \label{fig:amorization}
\end{figure*}

This introduces a \textit{generative} model for agents' behavior conditioned on their nested reasoning, $p(is_{i, \ell}^{1:T}, o_i^{1:T}, a_i^{1:T})$.

\textbf{Inference.} The level-$\ell$ agent $i$ performs inference to obtain the belief $b_{i, \ell}^t := p(is_{i, \ell}^t \given o_i^{1:t}, a_i^{1:t - 1})$.
$i$'s interactive state at level $\ell$ (i.e., $is_{i, \ell}^t = (s^t, b_{j, \ell - 1}^t, \theta_j)$) contains $j$'s belief at level $\ell - 1$ (i.e., $b_{j, \ell - 1}^t = p(is_{j, \ell - 1}^t \given o_j^{1:t}, a_j^{1:t - 1})$). Thus, inference at level $\ell$ depends on inference at level $\ell - 1$ which depends on inference at level $\ell - 2$, and so on. This recursion terminates at level $0$ in which the inference is the same as in a POMDP. That is, the belief is $b_{i, 0} = p(s^t \given o_i^{1:t}, a_i^{1:t - 1})$, disregarding the other agent.

\textbf{Planning.} The approximate planner of a level-$\ell$ agent $i$ determines its policy given its belief $b_{i, \ell}^t$ and $i$'s goal $\theta_i$. Inference in I-POMDP requires planning.
Given observations $o_i^{1:t}$ and previous actions $a_i^{1:t - 1}$, the full posterior under the generative model in \cref{fig:ipomdp} contains not only the interactive state $is_{i, \ell} = (s, b_{j, \ell - 1}, \theta_j)$ but also $j$'s actions and observations, $a_{j, \ell - 1}, o_{j, \ell - 1}$, which need to be marginalized out. Marginalizing actions requires evaluating $j$'s policy at level $\ell - 1$, $\pi_{j, \ell - 1}(a_{j, \ell - 1}^t \given b_{j, \ell - 1}^t, \theta_j)$. As illustrated in Figure~\ref{fig:ipomdp}, planning also requires inference. For example, agent $i$ needs to simulate how its own actions may change agent $j$'s belief in order to optimally interact with agent $j$.

\section{Amortized Inference for I-POMDPs}




Since planning and inference at level $\ell$ depend on the nested inference of beliefs at lower levels, it can become prohibitively expensive to reason about agents with high $\ell$.
Hence, we propose to \emph{amortize inference} at each level.
At level $\ell$, we learn a neural recognition model $q_\phi^\ell(is_{i, \ell}^{1:T} | o_i^{1:T}, a_i^{1:T})$ parameterized by $\phi$ to approximate $p(is_{i, \ell}^{1:T} | o_i^{1:T}, a_i^{1:T})$ by minimizing the KL divergence between exact inference and the recognition model on data sampled from the generative model $p(is_{i, \ell}^{1:T}, o_i^{1:T}, a_i^{1:T})$:
\begin{equation}
\mathcal L(\phi, \ell) = \E\left[\text{KL}(p(is_{i, \ell}^{1:T} | o_i^{1:T}, a_i^{1:T}) || q_\phi^\ell(is_{i, \ell}^{1:T} | o_i^{1:T}, a_i^{1:T}))\right].   
    \label{eq:amortized_inference_loss}
\end{equation}


Sampling at level $\ell$ requires inference at level $\ell - 1$. To accelerate the sampling, we amortize the lower level inference using a previously trained recognition model at that level, $q_\phi^{\ell - 1}(is_{j, \ell - 1}^{1:T} | o_j^{1:T}, a_j^{1:T})$. At level-0, we learn a recognition model over states $q_\phi^0(s_{1:T} | o_i^{1:T}, a_i^{1:T})$.



The algorithm for training a set of recognition models, $q_\phi^0, \dotsc, q_\phi^\ell$, is outlined as follows:
\begin{itemize}
    \item Train a recognition model $q_\phi^0(s_{1:T} | o_{1:T}, a_{1:T})$ by generating data from $p(s^{1:T}, o_i^{1:T}, a_i^{1:T})$ and minimizing $\mathcal L(\phi, 0)$.
    \item for levels $\ell = 1, \dotsc, L$
    \begin{itemize}
        \item Train $q_\phi^\ell(is_{i, \ell}^{1:T} | o_i^{1:T}, a_i^{1:T})$ by generating data from $p(is_{i, \ell}^{1:T}, o_i^{1:T}, a_i^{1:T})$ and minimizing $\mathcal L(\phi, \ell)$.
    \end{itemize}
\end{itemize}

We describe how we generate the training data in the supplementary material.

\textbf{Factorization of the recognition model.} We factorize the recognition model autoregressively:
\begin{align}
    q_\phi^\ell(is_{i, \ell}^{1:T} | o_i^{1:T}, a_i^{1:T})
    &= \prod_{t = 1}^T q_\phi^\ell(is_{i, \ell}^t | is_{i, \ell}^{t - 1}, o_i^{1:t}, a_i^{1:t - 1}).
\end{align}
This allows us to approximate the belief at any step $t$, $b_{i, \ell}^t = p(is_{i, \ell}^t \given o_i^{1:t}, a_i^{1:t - 1})$. We further factorize the recognition model at each step over beliefs, states, and goals:
\begin{align}
     q_\phi^\ell(is_{i, \ell}^t | is_{i, \ell}^{t - 1}, o_i^{1:t}, a_i^{1:t - 1}) 
     &=q_\phi^{\ell}(b_{j, \ell - 1}^t | is_{i, \ell}^{t - 1}, o_i^{1:t}, a_i^{1:t - 1}) \nonumber\\
     &\quad\cdot q_\phi(s^t | is_{i, \ell}^{t - 1}, o_i^{1:t}, a_i^{1:t - 1})\nonumber\\
     &\quad\cdot q_\phi^{\ell}(\theta_{j} | is_{i, \ell-1}^{t - 1}, o_i^{1:t}, a_i^{1:t - 1}).
\end{align}
One remaining challenge is parameterizing the distribution over beliefs at the lower level $\ell - 1$, $b_{j, \ell - 1}^t$. For this, we first examine the distribution over $b_{j, \ell - 1}^t$ under the \emph{prior}, $p(b_{j, \ell - 1}^t | o_{j, \ell - 1}^t, b_{j, \ell - 1}^{t - 1}, a_{j, \ell - 1}^{t - 1})$, following the factorization in \cref{fig:ipomdp}.
Under the prior, this distribution is a \emph{deterministic} belief-update function of the current observation, previous belief, and previous action. Since such belief updates are in general intractable, we represent beliefs as a set of $N$ weighted samples (or particles), $\{(is_{j, \ell - 1}^{t, (n)}, w_{j, \ell - 1}^{t, (n)})\}_{n = 1}^{N_\ell}$. We perform a particle update, $b_{j, \ell - 1}^t = \text{ParticleUpdate}(o_{j, \ell - 1}^t, b_{j, \ell - 1}^{t - 1}, a_{j, \ell - 1}^{t - 1})$, at each step.
In this paper, we set the recognition distribution over $b_{j, \ell - 1}^t$ to be \emph{identical to the prior}. That is, to sample from the recognition model $q_\phi(b_{j, \ell - 1}^t | o_{j, \ell - 1}^t, b_{j, \ell - 1}^{t - 1}, a_{j, \ell - 1}^{t - 1})$, we perform the particle update above.
By doing so, during importance sampling described below (Eq.~(\ref{eq:weight})), the ratio $p(b_{j, \ell - 1}^t | \cdots) / q_\phi(b_{j, \ell - 1}^t | \cdots)$ conveniently becomes one. 

In sum, we need to train a recognition model for state inference, $q_\phi(s^t|\cdots)$, shared by all levels; additionally, for \textit{each} level, we train a recognition model for goal inference at that level, $q_\phi^\ell(\theta_j^t|\cdots)$.

\textbf{Importance sampling.} As Figure~\ref{fig:amorization} illustrates, we sample $N_\ell$ particles based on the recognition networks at each level $\ell$ at time $t$. We can then compute the importance weight for a particle at level $\ell$ at time $t$ as


\begin{align}
    w_t &= \frac{p(is_{i, \ell}^{1:t}, o_i^{1:t} | a_i^{1:t - 1})}{q_\phi^\ell(is_{i, \ell}^{1:t} | o_i^{1:t}, a_i^{1:t - 1})} \nonumber\\
    &=\frac{\sum_{a_j^{1:t-1}}p(s^{1:t}| a_i^{1:t - 1},a_j^{1:t-1})\pi_{j,\ell-1}(a_j^{1:t-1} | b_{j,\ell-1}^{1:t-1},\theta_{j,\ell})}{q_\phi(s^{1:t} | o_i^{1:t}, a_i^{1:t - 1})q_\phi^\ell(\theta_{j} | o_i^{1:t}, a_i^{1:t - 1})}.\label{eq:weight}
\end{align}

We use this weight to refine the posterior approximated by the recognition model. We show the derivation of Eq.~(\ref{eq:weight}) in the supplementary material. The algorithm for approximating the belief $b_{i, \ell}^t$ using importance sampling using $N_\ell$ samples is
\begin{itemize}
    \item For sample $n = 1, \dotsc, N_\ell$
    \begin{itemize}
        \item For time $\tau = 1, \dotsc, t$
        \begin{itemize}
            \item Sample $is_{i, \ell}^{\tau, (n)} \sim q_\phi^\ell(\cdot | is_{i, \ell}^{\tau - 1, (n)}, o_i^{1:\tau}, a_i^{1:\tau - 1})$
        \end{itemize}
        \item Evaluate $w_t^{(n)}$ from \eqref{eq:weight}.
    \end{itemize}
\end{itemize}
The set of weighted samples $\{(is_{i, \ell}^{t, (n)}, w_t^{(n)})\}_{n = 1}^{N_\ell}$ is used to approximate the posterior belief $b_{i, \ell}^t$. To approximate a sequence of beliefs from 1 to $T$, we re-run this importance sampling procedure at every time step. We found that this performs better than sequential Monte Carlo because the quality of the samples from the recognition models drastically increases as the models observe more steps. Re-running importance sampling at every step ensures that our method fully utilizes the higher-quality samples produced by the recognition models at later steps. 

\section{Experiments}
\label{sec:experiments}
\subsection{Construction Environment}

\subsubsection{Setup}
\begin{figure}[t!]
\centering
\includegraphics[width=0.3\textwidth]{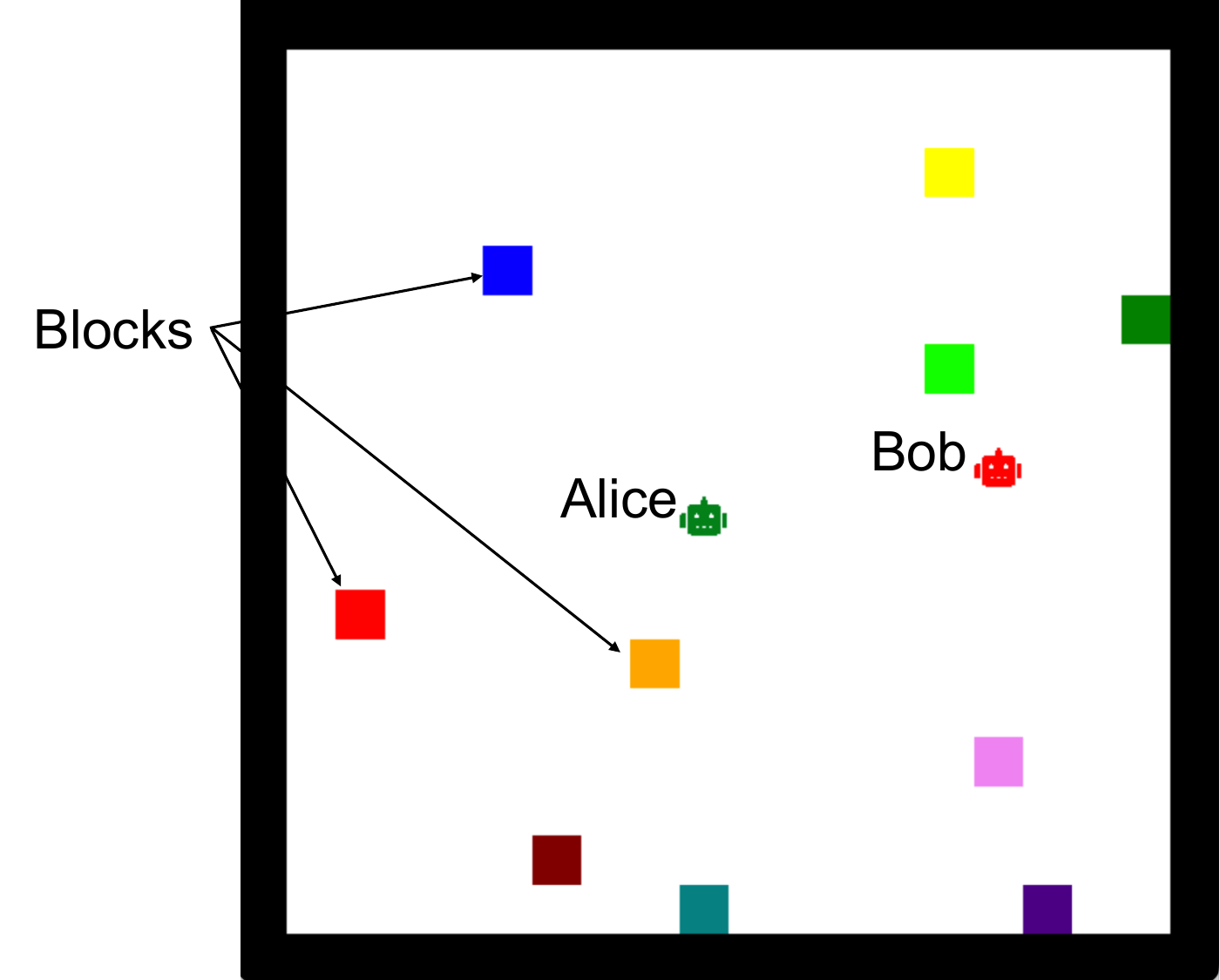}
    \caption{An example environment in \textbf{Construction}.
    }
    \label{fig:construction_setup}
\end{figure}

Inspired by \citet{ullman2009help} and \citet{tejwani2022social}, we evaluate our method in a 2D grid-world domain, \textbf{Construction}, as illustrated in Figure~\ref{fig:construction_setup}. There are two agents in this domain: \textit{Alice} (a level-0 agent) whose goal is to put two of the blocks next to each other and \textit{Bob} (a level-1 agent) whose goal is to either help or hinder Alice. Both agents can observe the full state and each other's actions but do not know each other's goals. Each agent can move in 4 directions. Once an agent is on top of a block, it automatically picks it up. When the agent is carrying a block, it can put down the block at any step. At each step, a model is asked to infer Bob's goal (helping or hindering) from a level-2 third-person observer's perspective, based on the observed actions of Alice and Bob up until that step, i.e., online inference of Bob's goal. Since Bob is also inferring Alice's goal in order to help or hinder Alice, a successful online goal inference must infer how Bob infers Alice's goal as well. In each episode, there are 10 blocks randomly placed in 20 $\times$ 20 grid, and the two agents are randomly spawned in the grid. There are 45 possible goals for Alice and 2 possible goals for Bob. The hypothesis space in this domain (90 hypotheses in total) is much larger than that of prior works. For instance, there are only 2 - 4 hypotheses in \citet{ullman2009help} and \citet{tejwani2022social}. 

\begin{figure*}[t!]
     \centering
         \includegraphics[width=1.0\textwidth]{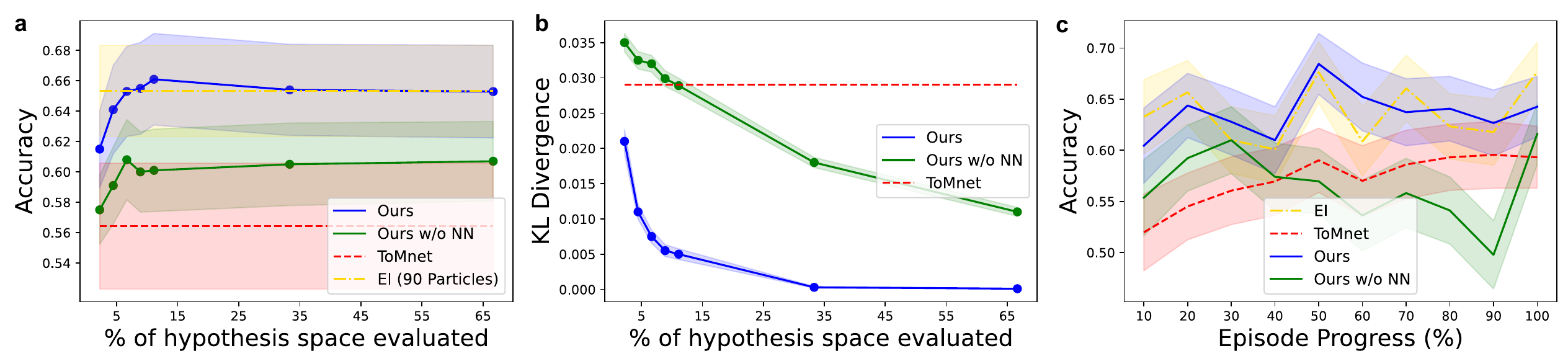}
         \caption{Online goal inference performance in \textbf{Construction}. (\textbf{a}) Accuracy of all steps over the number of particles (in terms of the percentage of all 90 hypotheses). (\textbf{b}) KL-divergence between model inference and exact inference. (\textbf{c}) Averaged goal inference accuracy over the progress of an episode. Ours and Ours w/o NN use 10 particles. The shaded areas show the standard errors of different methods. Note that we omit the standard error of ToMnet in (\textbf{b}) since it is too large for the figure (0.2).}
     \label{fig:construction_results}
\end{figure*}

\begin{figure*}[t!]
     \centering
         \centering
         \includegraphics[width=0.9\textwidth]{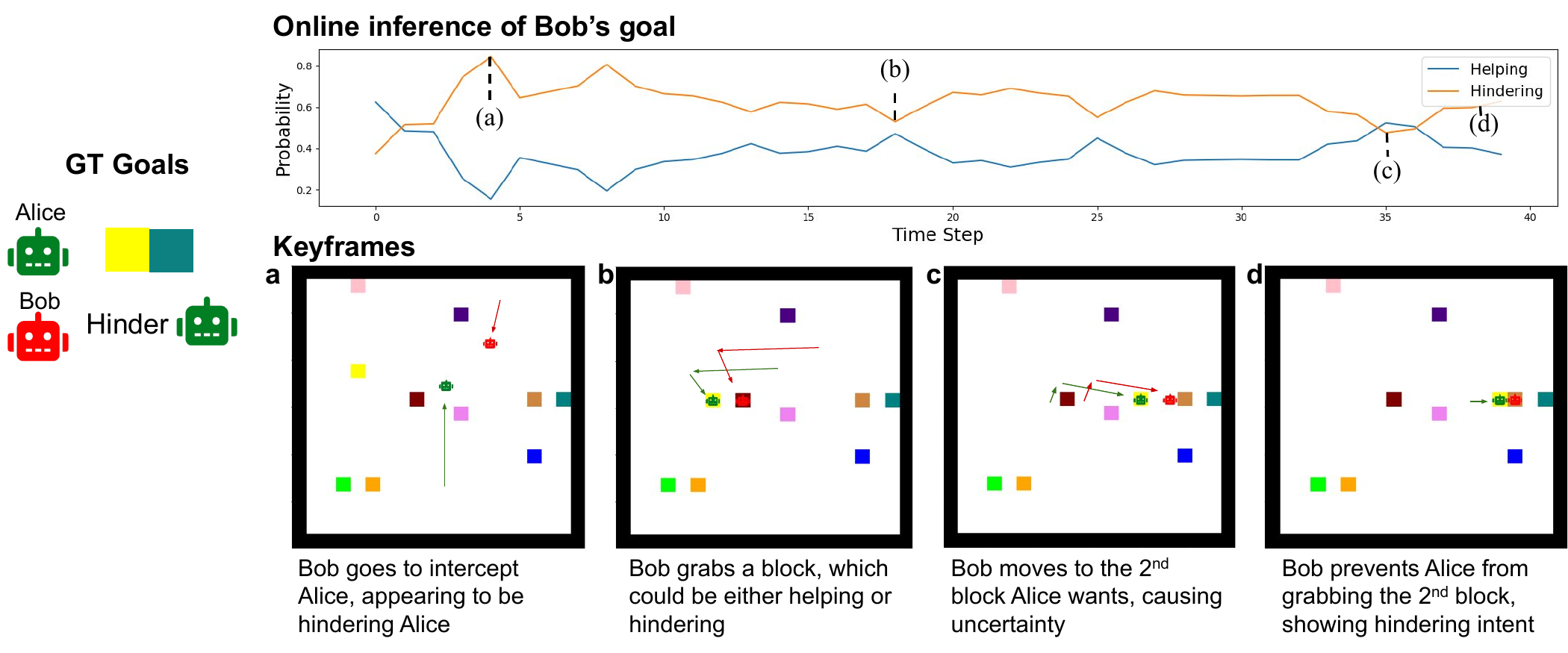}
        \caption{Online goal inference of our method (with 10 particles) in a typical episode in \textbf{Construction}, in which Alice wants to put yellow and teal blocks together and Bob tries to hinder Alice. The plot on the top shows the posterior probabilities of the two hypotheses based on our method's inference at any given time step. The keyframes on the bottom explain why our method adjusts its inference. Note that the arrows in the frames show the trajectories of the agents.}
        \label{fig:construction_example}
\end{figure*}

\subsubsection{Implementation}
We implement breath-first-search (BFS) for both agents' planners and use exact inference for I-POMDP (by enumerating all possible goal hypotheses) for both agents at each level to synthesize the actions of the agents. This allows us to create two training sets -- S1 with Alice acting alone and S2 with Bob interacting with Alice. Note that S1 and S2 correspond to the training set for level-1 inference and level-2 inference respectively. We also synthesize 100 testing episodes of Alice and Bob interacting with each other. We provide more details in the supplementary material.

\subsubsection{Results}

We compare our method (\textbf{Ours}) against the following baselines.

\noindent\textbf{ToMnet}: We adopt ToMnet proposed in \citet{rabinowitz2018machine} to infer Bob's goal. The network is trained with cross-entropy loss.

\noindent\textbf{Ours$_\text{-NN}$}: This baseline uses a uniform distribution as the proposal distribution instead of neural amortized inference.

We also show the exact inference (\textbf{EI}) performance, in which we enumerate all possible hypotheses to solve I-POMDP. Note that for Ours and Ours$_\text{-NN}$, we amortize the level-1 goal inference by sampling a small number of particles to propose possible goals for Alice and enumerate all 2 possible goals for level-2 goal inference (Bob's goal). 

We report the averaged goal inference accuracy over the number of sampled particles in terms of the percentage of the full hypothesis space (90 hypotheses of both agents' goals) in  Figure~\ref{fig:construction_results}\textbf{a}. From the results, we can see that our method's inference accuracy becomes comparable to the exact inference with only 6 particles ($6.7\%$ of the hypothesis space). It is much more efficient compared to the exact inference which needs 90 particles. With uniform proposals (Ours$_\text{-NN}$), on the other hand, the inference accuracy is much lower than the full model that utilizes the data-driven proposals from the recognition network. This demonstrates that neural amortized inference can sample high-quality hypotheses to drastically reduce the amount of computation necessary for producing accurate inference. Interestingly, the inference accuracy of the level-1 goal recognition network is only around 17\%. This suggests that we do not need to train a highly accurate network, as long as the proposal distribution is better than the uniform distribution. In contrast, ToMnet achieves poor accuracy, as it neither evaluates hypotheses with planning nor explicitly reasons about how the level-1 agent infers the level-0 agent's goals.

The accuracy of our method maintains at a similar level after using more than 6 particles but has a more accurate estimation of the uncertainty of the hypotheses. This is reflected in the lower KL divergence between our method's inference and the exact inference when there are more particles as shown in Figure~\ref{fig:construction_results}\textbf{b}. When considering more hypotheses, the model will likely discover alternative hypotheses that can explain the observed behavior equally well. 

Figure~\ref{fig:construction_results}\textbf{c} demonstrates how each method's online goal inference accuracy changes over time. Specifically, we plot the averaged accuracy across all testing episodes as a function of what percentage of an episode a method has seen. The result of Ours is based on 10 particles. The inference of all methods becomes more accurate as they observe more actions. However, the accuracy of ToMnet increases more slowly and reaches a lower plateau compared to EI and Ours.


Figure~\ref{fig:construction_example} illustrates a typical example of the online goal inference conducted by our method. It shows that our method not only correctly infers the goal, but can also adjust the certainty in inference dynamically by evaluating alternative hypotheses. For instance, when Bob grabs or moves toward the 2nd block (frames (\textbf{b}) and (\textbf{c}) in Figure~\ref{fig:construction_example}), he could try to help Alice by delivering the block to her; he could also try to hinder Alice by moving it away from her. At these moments, our model decreases its certainty. However, with further observation, our model gradually increases its confidence again as the behavior shows a clearer hindering intent. Such uncertainty estimation is key to the robustness of multi-agent nested reasoning, which can be achieved by only sampling a small number of particles with our method. We include additional results in the supplementary material.

\subsection{Driving Environment}

\subsubsection{Setup} 
\begin{figure}[t!]
\centering
\includegraphics[width=0.35\textwidth]{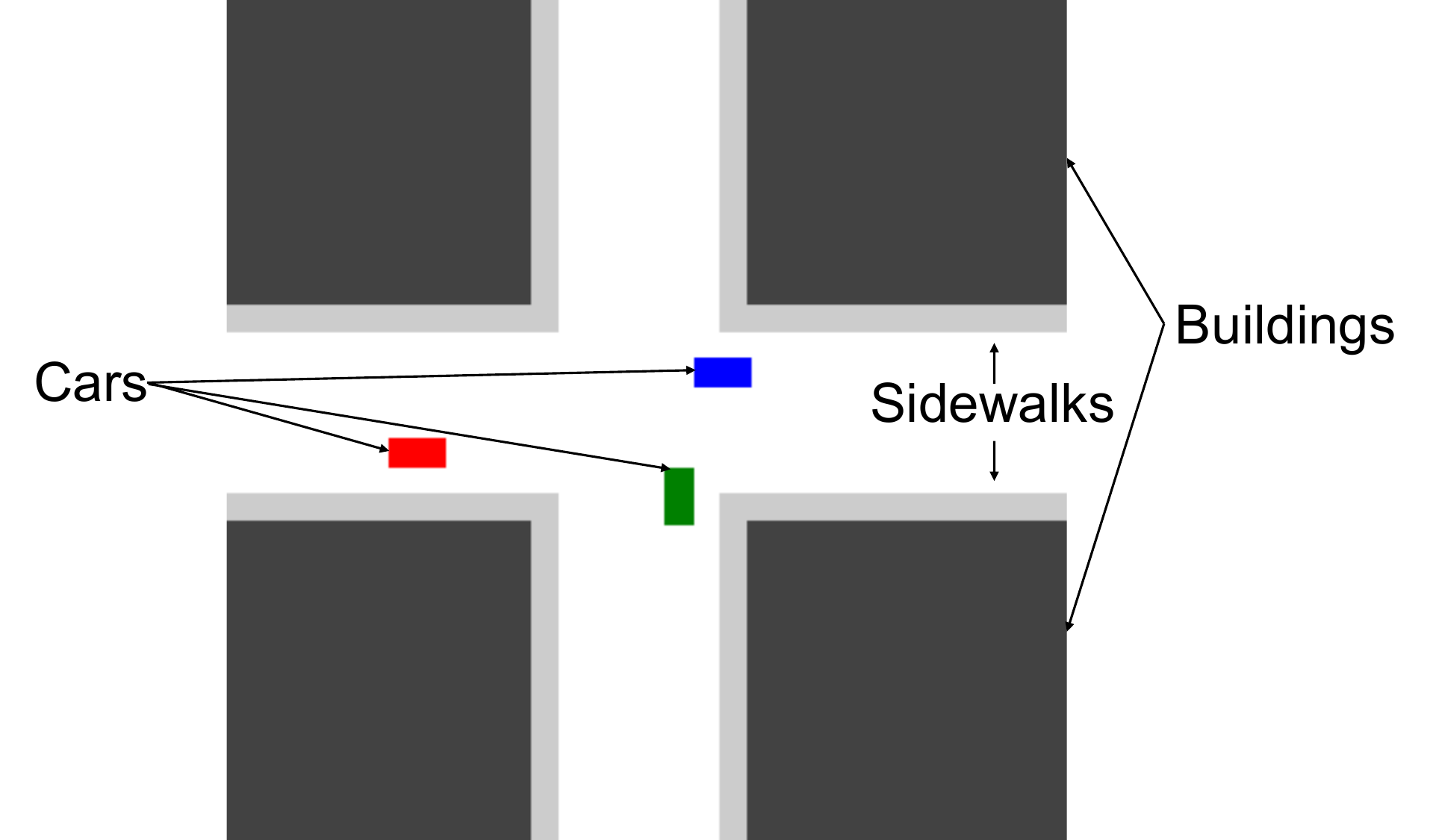}
    \caption{An example environment in \textbf{Driving}. The dark blocks are buildings and the light gray regions are sidewalks.
    }
    \label{fig:driving_setup}
\end{figure}

For the second experiment, we simulate the traffic at an intersection as shown in Figure~\ref{fig:driving_setup} using a commonly used driving environment, CARLO \citep{cao2020reinforcement}. In this environment, we randomly assign a goal (forward, left-turn, or right-turn) to a driver, indicating the destination after the intersection. A car can be controlled through 5 actions at each step: accelerating, braking, rotating left, rotating right, and signaling danger to other drivers by honking. Each driver has only a partial observation of the world. They cannot see cars out of their fields of view or obstructed by other cars or buildings. To avoid crashing into other cars, a driver must infer other drivers' mental states (goals and beliefs about the states) recursively and consequently predict other drivers' future actions. When a driver infers that another driver was unaware of a nearby car, it is also necessary to signal danger by honking to avoid a potential crash. The task for a model is to predict the next action of a car from a third-person observer's perspective, which requires a level-2 inference in this environment. Unlike \textbf{Construction}, the nested reasoning in this domain includes both the inference of goals and the beliefs about the states.

\begin{figure*}[t!]
     \centering
         \includegraphics[width=1.0\textwidth]{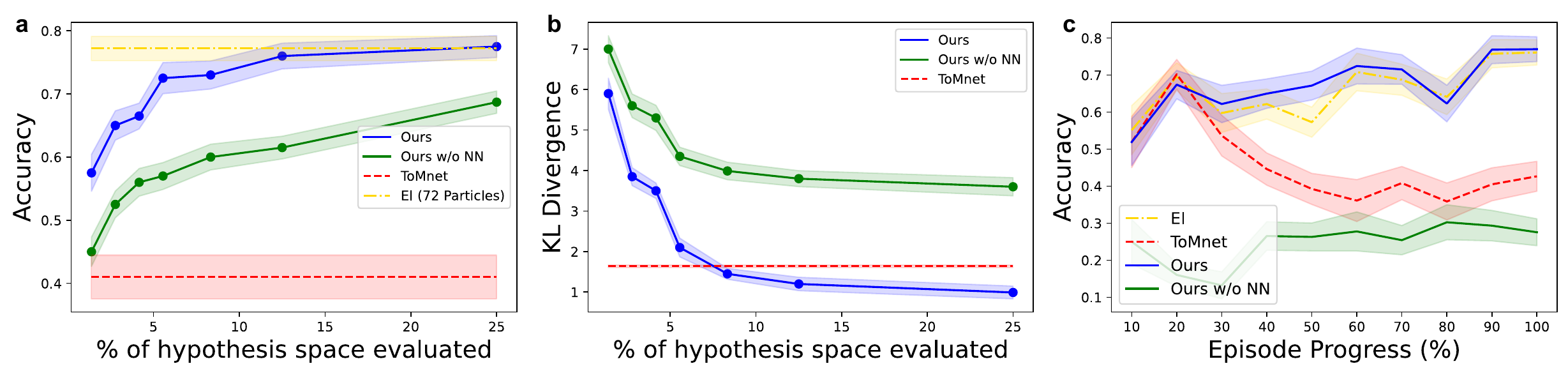}
         \caption{Action prediction performance in \textbf{Driving}. (\textbf{a}) Action prediction accuracy in a 3-car scenario. (\textbf{b}) KL-divergence between model inference and exact inference. (\textbf{c}) Averaged action prediction accuracy over the progress of an episode. Ours and Ours w/o NN all use 9 particles.}
     \label{fig:3 car normal metrics}
\end{figure*}

\begin{figure*}[t!]
     \centering
         \includegraphics[width=1.0\textwidth]{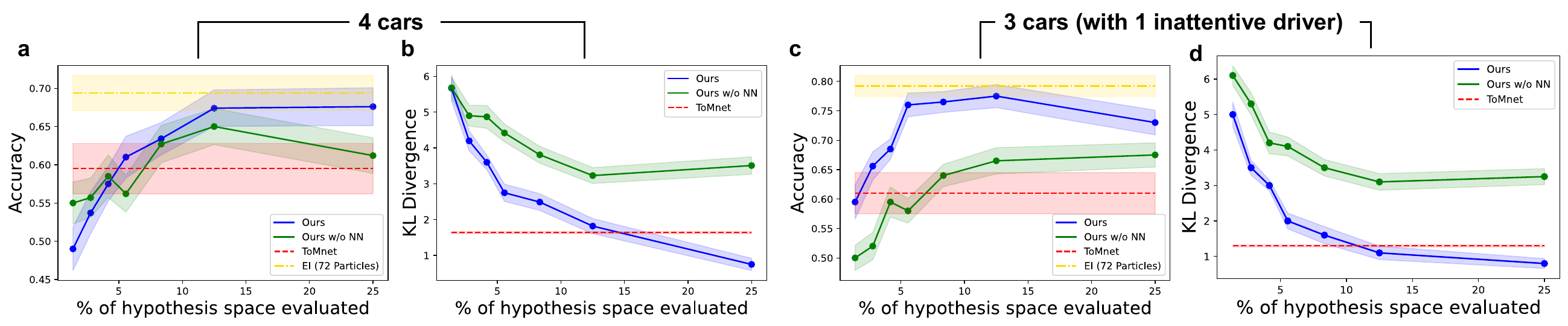}
         \caption{Generalization evaluation results of models trained with 3 cars controlled by normal drivers. (\textbf{a})(\textbf{b}) Results on episodes with 4 cars. (\textbf{c})(\textbf{d}) Results on episodes with 3 cars where one of the drivers is inattentive.}
     \label{fig:car_generalization}
\end{figure*}

\begin{figure}[t!]
     \centering
         \centering
         \includegraphics[width=0.47\textwidth]{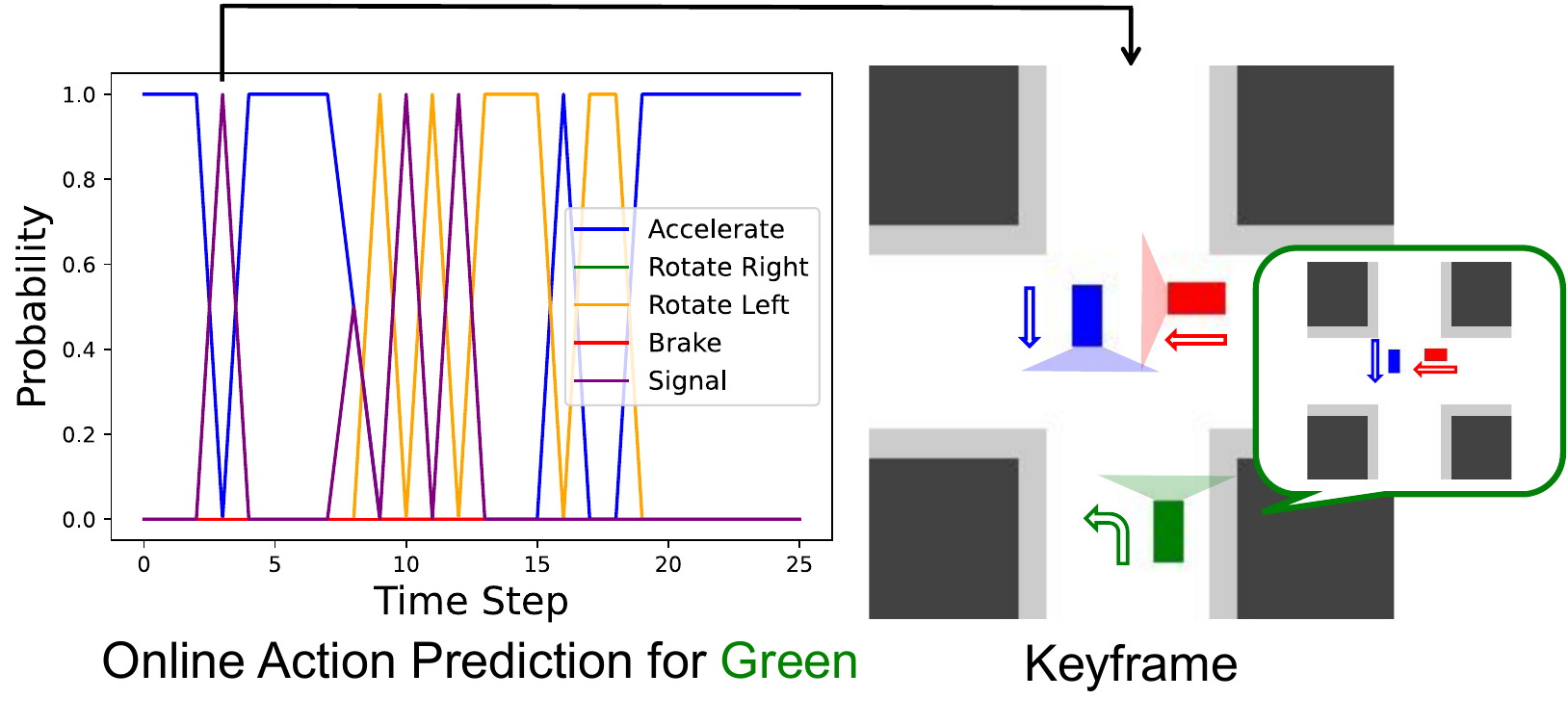}
        \caption{Online action prediction of our method (with 9 particles) in a typical episode in \textbf{Driving}. The plot on the left shows our method's prediction of green's action. The keyframe on the right explains why our method predicts that green will signal danger at the highlighted step. Note that the colored cones show the fields of view of the drivers and the arrows show the predicted goals. At this moment, the model infers that i) green wants to turn left and that ii) green infers that red wants to go forward and is unaware of green (as shown in the smaller frame). Thus our method predicts that green will signal danger to alert the red car to green's presence. The ground-truth action is indeed signaling danger.}
        \label{fig:car_example}
\end{figure}

\subsubsection{Implementation}

For each driver's belief, we represent the world state as the states of the cars in all 8 lanes. For the $k$-th lane, we model up to 2 cars that are closest to the intersection: $\{\langle e_{k,m}, (x_{k,m}, y_{k,m}), (\sigma_{k,m}, v_{k,m})\rangle\}_{m=1,2}$. The indices of cars, $m$, are ordered by their distances to the intersection. $e_{k,m}\in\{0,1\}$ indicates whether a car exists in the $k$-th lane. If a car exists ($e_{k,m} = 1$), $(x_{k,} \in \mathbb{R}, y_{k,m} \in \mathbb{R})$ indicates its location, we use $\sigma_{k,m} \in [-\pi,\pi]$ to indicate its heading angle and $v_{k,m} \in [0, v_\text{max}])$ to indicate its speed ($v_\text{max}$ is the maximum speed). 


All drivers' policies are based on a hierarchical planner. The planner first decides whether to move, stop or signal danger at each step; if it decides to move, it then selects the action (accelerate, rotate left or rotate right) that follows the shortest path; if it decides to stop, then it will brake; otherwise, it will signal danger by honking. We use exact inference for I-POMDP to synthesize driver behavior and create three training sets: S0 with level-0 drivers using exact state inference (for training the state recognition model); S1 with level-0 drivers using amortized state inference (for training the level-1 goal recognition model); and S2 with level-1 drivers using amortized level-1 inference (for training the level-2 goal recognition model). In all training sets, there are 3 cars. For evaluation, we synthesize 100 testing episodes of 3 cars interacting with each other. Additionally, we create two generalization testing sets: i) 4 cars and ii) 3 cars where one of the cars is controlled by an inattentive driver who is not paying attention to other cars. Each set has 100 episodes. To predict a driver's actions, we use level-1 agent policy, i.e., $\pi_{i, 1}(a_i^t | b_{i,1}^t, \theta_i)$. We provide more implementation details in the supplementary material.

\subsubsection{Results}

Similar to \textbf{Construction}, we compare our method (Ours) against EI, ToMnet, and Ours$_\text{-NN}$. Note that ToMnet here is trained to predict the action of each car.

From Figure \ref{fig:3 car normal metrics}, we can see that our method significantly outperforms baselines. With only 9 particles ($12.5\%$ of the hypothesis space), our method's action prediction accuracy is already comparable to the exact inference (which requires 72 particles\footnote{We explain why it needs 72 particles in the supplementary material.}). Interestingly, unlike our method and the exact inference, ToMnet's accuracy drops drastically over time in (Figure \ref{fig:3 car normal metrics}\textbf{c}). This is because later in an episode, cars are closer to each other and thus have to carefully coordinate with each other based on nested reasoning. This further demonstrates the difficulty of this domain and the necessity of robust nested reasoning in understanding and predicting complex multi-agent interactions.

We further evaluate the generalizability of different methods. In particular, all methods are trained using episodes with 3 cars controlled by normal drivers. We test the methods on episodes with 4 cars (Figure~\ref{fig:car_generalization}\textbf{ab}) and on episodes with 3 cars in which one of the drivers is inattentive (Figure~\ref{fig:car_generalization}\textbf{cd}). In both cases, our method still performs significantly better than baselines. In Figure~\ref{fig:car_generalization}\textbf{c}, our method's accuracy drops a bit when using $25\%$ particles. This is because it estimates more hypotheses and consequently decreases its confidence in prediction. However, the KL-divergence becomes lower when the model uses more particles (Figure~\ref{fig:car_generalization}\textbf{d}), making the uncertainty estimation more accurate.


Figure~\ref{fig:car_example} depicts a typical example of online action prediction by our method (with 9 particles) in the 3 normal cars condition. By inferring how green infers red's goal and belief, our method is able to correctly anticipate green's action (i.e., signaling danger). Such prediction is made possible by sophisticated nested reasoning. In fact, due to the lack of robust nested reasoning, ToMnet consistently fails to predict any ``signal'' action correctly. We include additional examples in the supplementary material.


\section{Conclusion}
\label{sec:conclusion}
In this work, we propose a neural amortized inference approach to accelerate nested multi-agent reasoning. We evaluate our method in two complex multi-agent domains with large hypothesis spaces. The results demonstrate that our method can significantly improve the efficiency of nested reasoning while maintaining a high level of accuracy. In addition, our method can also estimate the uncertainty in its inference and generalize well to unseen scenarios. 

In the current experiments, we only amortize up to level-2 reasoning. However, our method is general enough to amortize higher-order reasoning as well, which we intend to evaluate in future work. We also plan to study how to infer the minimum level required to understand multi-agent interaction in a given domain. One possibility is to train a network to amortize the inference of the necessary level through meta-learning.

\section*{Acknowledgements}
This work was supported by the DARPA Machine Common Sense program, ONR MURI N00014-13-1-0333, and a grant from Lockheed Martin.

\bibliography{aaai24}

\end{document}